\documentclass[graybox]{svmult}

\usepackage{mathptmx}       
\usepackage{helvet}         
\usepackage{courier}        
\usepackage{makeidx}         
\usepackage{graphicx}        
\usepackage{multicol}        
\usepackage[bottom]{footmisc}
\usepackage[utf8]{inputenc}
\usepackage{hyperref}
\usepackage[ruled,vlined]{algorithm2e}
\usepackage[caption=false]{subfig}
\usepackage{wrapfig}

\makeindex             

\begin{document}

\title*{Incremental Improvement of a Question Answering System by Re-ranking Answer Candidates using Machine Learning}
\titlerunning{Incremental Improvement of a Question Answering System by Re-ranking}
\author{Michael Barz and Daniel Sonntag}
\institute{Michael Barz \at German Research Center for Artificial Intelligence, Saarland Informatics Campus, D3 2, 66123 Saarbrücken, Germany \email{michael.barz@dfki.de} \at Saarbrücken Graduade School of Computer Science
\and Daniel Sonntag \at German Research Center for Artificial Intelligence, Saarland Informatics Campus, D3 2, 66123 Saarbrücken, Germany \email{daniel.sonntag@dfki.de}}
%
%
\maketitle

\abstract*{
	We implement a method for re-ranking top-10 results of a state-of-the-art question answering (QA) system.
	The goal of our re-ranking approach is to improve the answer selection given the user question and the top-10 candidates.
	We focus on improving deployed QA systems that do not allow re-training or re-training comes at a high cost.
	Our re-ranking approach learns a similarity function using n-gram based features using the query, the answer and the initial system confidence as input.
	Our contributions are:
	(1) we generate a QA training corpus starting from 877 answers from the customer care domain of T-Mobile Austria, 
	(2) we implement a state-of-the-art QA pipeline using neural sentence embeddings that encode queries in the same space than the answer index, and
	(3) we evaluate the QA pipeline and our re-ranking approach using a separately provided test set.
	The test set can be considered to be available after deployment of the system, e.g., based on feedback of users.
	Our results show that the system performance, in terms of top-n accuracy and the mean reciprocal rank, benefits from re-ranking using gradient boosted regression trees. On average, the mean reciprocal rank improves by $9.15\%$.}

\abstract{We implement a method for re-ranking top-10 results of a state-of-the-art question answering (QA) system.
	The goal of our re-ranking approach is to improve the answer selection given the user question and the top-10 candidates.
	We focus on improving deployed QA systems that do not allow re-training or when re-training comes at a high cost.
	Our re-ranking approach learns a similarity function using n-gram based features using the query, the answer and the initial system confidence as input.
	Our contributions are:
	(1) we generate a QA training corpus starting from 877 answers from the customer care domain of T-Mobile Austria, 
	(2) we implement a state-of-the-art QA pipeline using neural sentence embeddings that encode queries in the same space than the answer index, and
	(3) we evaluate the QA pipeline and our re-ranking approach using a separately provided test set.
	The test set can be considered to be available after deployment of the system, e.g., based on feedback of users.
	Our results show that the system performance, in terms of top-n accuracy and the mean reciprocal rank, benefits from re-ranking using gradient boosted regression trees. On average, the mean reciprocal rank improves by $9.15\%$.}

\section{Introduction}
In this work, we examine the problem of incrementally improving deployed QA systems in an industrial setting. We consider the domain of customer care of a wireless network provider and focus on answering frequent questions (focussing on the long tail of the question distribution \cite{Bernstein2012}).
In this setting, the most frequent topics are covered by a separate industry-standard chatbot based on hand-crafted rules by dialogue engineers.
Our proposed process is based on the augmented cross-industry standard process for data mining \cite{Sonntag2008} (augmented CRISP data mining cycle).
In particular, we are interested in methods for improving a model after its deployment through re-ranking of the initial ranking results. In advance, we follow the steps of the CRISP cycle towards deployment for generating a state-of-the-art baseline QA model. 
First, we examine existing data (\emph{data understanding}) and prepare a corpus for training (\emph{data preparation}). 
Second, we implement and train a QA pipeline using state-of-the-art open source components (\emph{modelling}). We perform an evaluation using different amounts of data and different pipeline configurations (\emph{evaluation}), also to understand the nature of the data and the application (\emph{business understanding}).
Third, we investigate the effectiveness and efficiency of re-ranking in improving our QA pipeline after the \emph{deployment} phase of CRISP. Adaptivity after deployment is modelled as \emph{(automatic) operationalisation} step with external reflection based on, e.g., user feedback. This could be replaced by introspective meta-models that allow the system to enhance itself by metacognition \cite{Sonntag2008}.
The QA system and the re-ranking approach are evaluated using a separate test set that maps actual user queries from a chat-log to answers of the QA corpus. Sample queries from the evaluation set with one correct and one incorrect sample are shown in Table \ref{tab:qa_samples}.

With this work, we want to answer the question whether a deployed QA system that is difficult to adapt and that provides a top-10 ranking of answer candidates, can be improved by an additional re-ranking step that corresponds to the operationalisation step of the augmented CRISP cycle. It is also important to know the potential gain and the limitations of such a method that works on top of an existing system.
We hypothesise that our proposed re-ranking approach can effectively improve ranking-based QA systems.

\begin{table}
	\centering
	\caption{Sample queries with a correct and an incorrect answer option according to our test set. We report the answers' rank of the baseline model that we used for our re-ranking experiments.}
	\label{tab:qa_samples}
	\small
	\begin{tabular}{p{3.5cm}|p{4.25cm}|p{3.75cm}}
		\textbf{User Query}	& \textbf{Correct Answer} & \textbf{Incorrect Answer} \\
		\hline
		Bekomme ich bei Vertragsverlängerung ein neues Handy? 
		(\emph{Do I get a new phone when extending my contract?})
		& Ab wann Sie Ihre Rufnummern verlängern können und welche Angebote bei einer Vertragsverlängerung auf Sie warten, sehen Sie in Ihrem persönlichen Kundenportal Mein T-Mobile [...]
		(\emph{In your online customer area of T-Mobile, you can see when you can continue your telephone numbers and which offers await you after extending your contract [...]}) \textbf{(rank 1)}
		& Suchen Sie ein neues Gerät, das genau Ihre Bedürfnisse und Anforderungen erfüllt? Sie wollen rechtzeitig über Neuerungen informiert werden? [...]
		(\emph{You are looking for a new device that satisfies all your requirements? You want to get recent news? [...]})
		\textbf{(rank 5)}  \\
		\hline
		tarife ohne bi\textbf{m}dung
		(\emph{plans without bi\textbf{m}ding contract}) -- \texttt{misspelled}
		& Wenn Sie bereits ein Handy besitzen und nur eine Simkarte benötigen, haben wir genau das Richtige für Sie: die Klax-SIM. [...]
		(\emph{If you own a new phone and all you need is a SIM card, we got exactly the right offer: the Klax-SIM. [...]})
		\textbf{(rank 3)}
		& Eine Übersicht über unsere aktuellen My Mobile Handytarife inklusive aller wichtigen Details finden Sie auf der Tarifseite. [...]
		(\emph{An overview of our current service plans with all important details can be found on our website. [...]})
		\textbf{(rank 1)} \\
		\hline
		
		Kosten für vertragübernahme
		(\emph{costs for a contract transfer})
		& Sie können Verträge an andere Personen übergeben, zusammenlegen oder trennen. Die Kosten belaufen sich auf [...] Ausführliche Informationen zum Thema finden Sie in den FAQ. 
		(\emph{You can transfer, join and split contracts from and to other persons. The costs are [...] More detailed information can be found in our FAQ.})
		\textbf{(rank 10)}
		& Ein Zukauf von Freiminuten ist nicht möglich und bei unseren aktuellen Tarifen auch nicht notwendig, da Freiminuten hier unlimitiert sind.
		(\emph{You cannot buy additional minutes. However, that's not required with our plans, because minutes are unlimited.})
		\textbf{(rank 1)} \\
		\hline
		
		Kreditkarte
		(\emph{credit card})
		& Eine Zahlung mittels Kreditkarte ist selbstverständlich bei uns möglich. Sollten Sie Ihre Zahlungsart auf Kreditkarte ändern oder Ihre Daten aktualisieren wollen, können Sie dies direkt über unseren LiveChat veranlassen. [...] 
		(\emph{Of course, you can pay with your credit card. If you want to change your payment settings to credit card or if you want to update your data, you can do so using our LiveChat. [...]})
		\textbf{(rank 7)}
		& Die Änderung Ihrer Kreditkartendaten ist zu Ihrer Sicherheit nur telefonisch bei der Serviceline unter [...] 
		(\emph{For security reasons, you can change your credit card data via phone using our service hotline at [...] only})
		\textbf{(rank 1)} \\
		\hline
		
		Hallo, ich möchte ein iPhone 7 kaufen (Ratenzahlung). Ich hab schon ein Vertrag (bis 09.2017)..wenn ich das verlängern möchte muss ich die Raten von meine altes Handy weiter zahlen? Lg
		(\emph{Hello, I'd like to buy an iPhone 7 (paying by instalments). I have a contract (till 09/2017)..if I want to extend it, do I need to pay the remaining rates for my old phone? Kind regards})
		& Ratenzahlungen oder Stundungen bei offenen Rechnungsbeträgen bietet T-Mobile NICHT an [...]
		(\emph{T-Mobile does NOT offer payment by instalments or deferred payments for outstanding bill amounts.})
		\textbf{(not in top-10)}
		& Mit der Umstellung auf LTE hat sich nichts am Geschwindigkeitsprofil inklusive der erreichbaren Maximalgeschwindigkeiten Ihres aktuellen Tarifes geändert. [...] 
		(\emph{The transition to LTE (4G) does not affect the maximum data transfer rate of your present service plan.})
		\textbf{(rank 1)} \\
		\hline
		
	\end{tabular}
\end{table}


\section{Related Work}
The broad field of QA includes research ranging from retrieval-based \cite{Xue2008,Das2016,Minaee2017,N18-2092} to generative \cite{Serban2015,Serban2015a}, as well as, from closed-domain \cite{eric-manning:2017:SIGDIAL,Oraby2017} to open-domain QA \cite{Serban2015a,Joshi2017,D16-1264,P17-1171}.
We focus on the notion of improving an already deployed system. 

For QA dialogues based on structured knowledge representations, this can be achieved by maintaining and adapting the knowledgebase \cite{SonntagEHPPRR07,Sonntag09,Sonntag2010_book}.
In addition, \cite{Sonntag2008} proposes metacognition models for building self-reflective and adaptive AI systems, e.g., dialogue systems, that improve by introspection.
Buck et al. present a method for reformulating user questions: their method automatically adapts user queries with the goal to improve the answer selection of an existing QA model \cite{DBLP:journals/corr/BuckBCGHGW17}.

Other works suggest humans-in-the-loop for improving QA systems.
Savenkov and Agichtein use crowdsourcing for re-ranking retrieved answer candidates in a real-time QA framework \cite{Savenkov2016a}.
In Guardian, crowdworkers prepare a dialogue system based on a certain web API and, after deployment, manage actual conversations with users~\cite{Huang2015}.
EVORUS learns to select answers from multiple chatbots via crowdsourcing \cite{Huang2018}. The result is a chatbot ensemble excels the performance of each individual chatbot.
Williams et al. present a dialogue architecture that continuously learns from user interaction and feedback \cite{Williams2017}.

We propose a re-ranking algorithm similar to \cite{Savenkov2016a}: we train a similarity model using n-gram based features of QA pairs for improving the answer selection of a retrieval-based QA system.

\section{Question Answering System}

\begin{figure}
	\sidecaption[t]
	\includegraphics[width=.4\columnwidth]{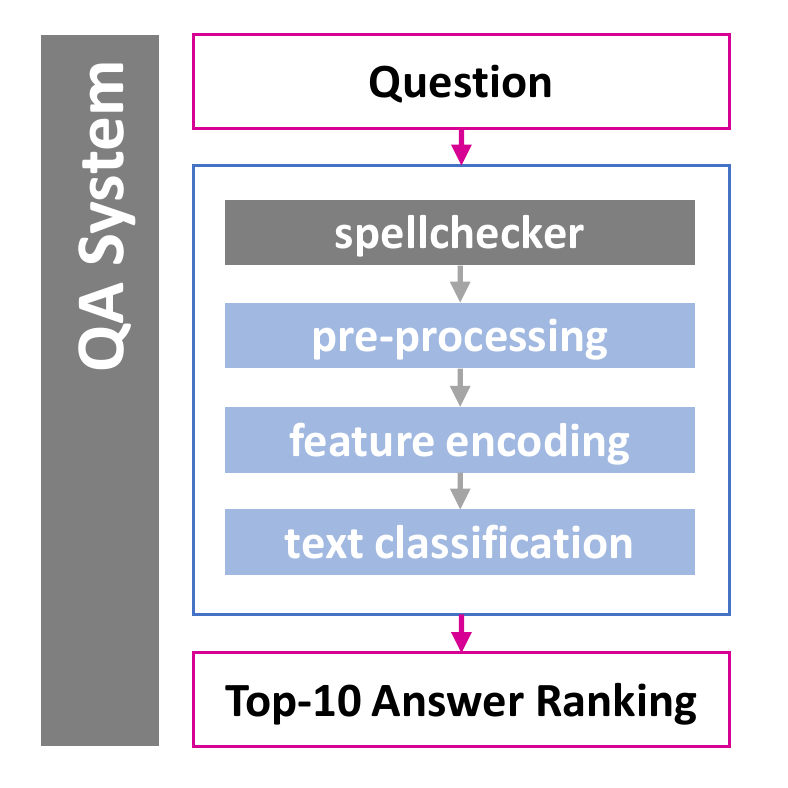}
	\caption{The basic configuration of the QA pipeline, which is a part of our complete QA system architecture with the re-ranking algorithm.}
	\label{fig:engine}
\end{figure}

We implement our question answering system using state-of-the-art open source components. Our pipeline is based on the Rasa natural language understanding (NLU) framework \cite{Bocklisch2017} which offers two standard pipelines for text classification: \emph{spacy\_sklearn} and \emph{tensorflow\_embedding}.
The main difference is that \emph{spacy\_sklearn} uses Spacy\footnote{\url{https://spacy.io/}} for feature extraction with pre-trained word embedding models and Scikit-learn \cite{scikit-learn} for text classification. In contrast, the \emph{tensorflow\_embedding} pipeline trains custom word embeddings for text similarity estimation using TensorFlow \cite{tensorflow2015-whitepaper} as machine learning backend. 
Figure \ref{fig:engine} shows the general structure of both pipelines.
We train QA models using both pipelines with the pre-defined set of hyper-parameters. For \emph{tensorflow\_embedding}, we additionally monitor changes in system performance using different epoch configurations\footnote{\url{https://rasa.com/docs/nlu/components/\#intent-classifier-tensorflow-embedding}}.
Further, we compare the performances of pipelines with or without a spellchecker and investigate whether model training benefits from additional user examples by training models with the three different versions of our training corpus including no additional samples (\texttt{kw}), samples from 1 user (\texttt{kw+1u}) or samples from 2 users (\texttt{kw+2u}) (see section Corpora). All training conditions are summarized in Table \ref{tab:qa_pipeline_configs}. 
Next, we describe the implementation details of our QA system as shown in Figure \ref{fig:engine}: the spellchecker module, the subsequent pre-processing and feature encoding, and the text classification. We include descriptions for both pipelines.

\begin{table}
	\centering
	\caption{Considered configurations for QA pipeline training.}
	\label{tab:qa_pipeline_configs}
	\begin{tabular}{r|c|p{3cm}|}
		& \emph{spacy\_sklearn} & \emph{tensorflow\_embedding} \\
		\hline
		parameters	& default & default with \texttt{epochs} $\in\{10,50,100,300,600\}$ \\
		\hline
		spellchecking 		&\multicolumn{2}{c|}{\texttt{yes}, \texttt{no}}  \\
		\hline
		training corpus	&\multicolumn{2}{c|}{\texttt{kw}, \texttt{kw+1u}, \texttt{kw+2u}}\\
		\hline
	\end{tabular}
\end{table}

\runinhead{Spellchecker}
We address the problem of frequent spelling mistakes in user queries by implementing an automated spell-checking and correction module. It is based on a Python port\footnote{\url{https://github.com/mammothb/symspellpy}} of the SymSpell algorithm\footnote{\url{https://github.com/wolfgarbe/SymSpell}} initialized with word frequencies for German\footnote{German 50k: \url{https://github.com/hermitdave/FrequencyWords}}. We apply the spellchecker as first component in our pipeline.

\runinhead{Pre-Processing and Feature Encoding.}
The \emph{spacy\_sklearn} pipeline uses Spacy for pre-processing and feature encoding. Pre-processing includes the generation of a Spacy \texttt{document} and tokenization using their German language model \linebreak \texttt{de\_core\_news\_sm} (v2.0.0).
The feature encoding is obtained via the \texttt{vector} function of the Spacy document that returns the mean word embedding of all tokens in a query.
For German, Spacy provides only a simple dense encoding of queries (no proper word embedding model). 

The pre-processing step of the \emph{tensorflow\_embedding} pipeline uses a simple whitespace tokenizer for token extraction. The tokens are used for the feature encoding step that is based on Scikit-learn's \texttt{CountVectorizer}. It returns a bag of words histogram with words being the tokens (1-grams).

\runinhead{Text Classification.}
The \emph{spacy\_sklearn} pipeline relies on Scikit-learn for text classification using a support vector classifier (SVC). The model confidences are used for ranking all answer candidates; the top-10 results are returned.

Text classification for \emph{tensorflow\_embedding} is done using TensorFlow with an implementation of the StarSpace algorithm \cite{DBLP:journals/corr/abs-1709-03856}. This component learns (and later applies) one embedding model for user queries and one for the answer id. It minimizes the distance between embeddings of QA training samples.
The distances between a query and all answer ids are used for ranking.

\subsection{Corpora}
In this work, we include two corpora: one for training the baseline system and another for evaluating the performance of the QA pipeline and our re-ranking approach. In the following, we describe the creation of the training corpus and the structure of the test corpus. Both corpora have been anonymised.

\runinhead{Training Corpus.}
The customer care department provides $877$ answers to common user questions.
Each answer is tagged with a variable amount of keywords or key-phrases ($M=3.81$, $SD=5.92$), $3338$ in total. 
We asked students to augment the training corpus with, in total, two additional natural example queries. This process can be scaled by crowdsourcing for an application in productive systems that might include more answers or that requires more sample question per answer or both.
The full dataset contains, on average, $5.81$ sample queries per answer totalling in $5092$ queries overall.
For model training, all questions (including keywords) are used as input with the corresponding answer as output.
We generated three versions of the training corpus: keywords only (\texttt{kw}, $n=3338$), keywords with samples from 1 user (\texttt{kw+1u}, $n=4215$) and keywords with samples from 2 users (\texttt{kw+2u}, $n=5092$).

\runinhead{Evaluation Corpus.}

The performance of the implemented QA system and of our re-ranking approach is assessed using a separate test corpus. It includes $3084$ real user requests from a chat-log of T-Mobile Austria, which are assigned to suitable answers from the training corpus (at most three). The assignment was performed manually by domain experts of the wireless network provider.
We use this corpus for estimating the baseline performance of the QA pipeline using different pipeline configurations and different versions of the training corpus.
In addition, we use the corpus for evaluating our re-ranking approach per cross-validation: we regard the expert annotations as offline human feedback.
The queries in this corpus contain a lot of spelling mistakes. We address this in our QA pipeline generation by implementing a custom spell-checking component.

\section{Baseline Performance Evaluation}

We evaluate the baseline model using all training configurations in Table \ref{tab:qa_pipeline_configs} to find a well-performing baseline for our re-ranking experiment. We use the evaluation corpus as reference data and report the top-1 to top-10 accuracies and the mean reciprocal rank for the top-10 results (MRR@10) as performance metrics.
For computing the top-n accuracy, we count all queries for which the QA pipeline contains a correct answer on rank 1 to n and divide the result by the number of test queries.
The MRR is computed as the mean of reciprocal ranks over all test queries. The reciprocal rank for one query is defined as $RR=\frac{1}{rank}$: The RR is $1$ if the correct answer is ranked first, $0.5$ if it is at the second rank and so on. We set RR to zero, if the answer is not contained in the top-10 results.

\begin{figure}
	\sidecaption[t]
	\includegraphics[width=.64\textwidth]{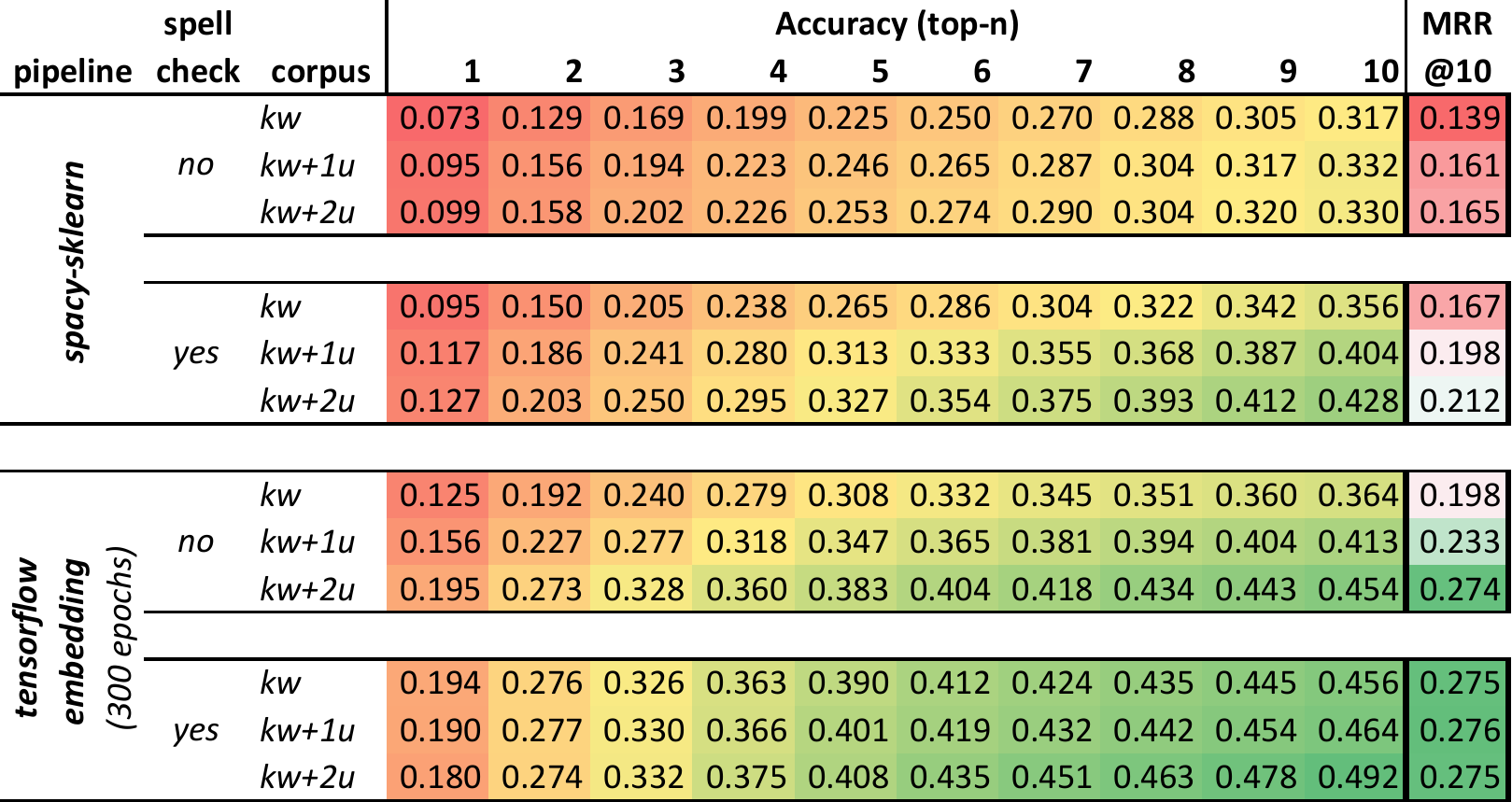}
	\caption{Performance metrics in terms of top-1 to top-10 accuracy and MRR@10 of both QA pipelines for different pipeline configurations and training corpora.}
	\label{fig:res_pipeline}
\end{figure}

\begin{figure}
	\sidecaption[t]
	\includegraphics[width=.59\textwidth]{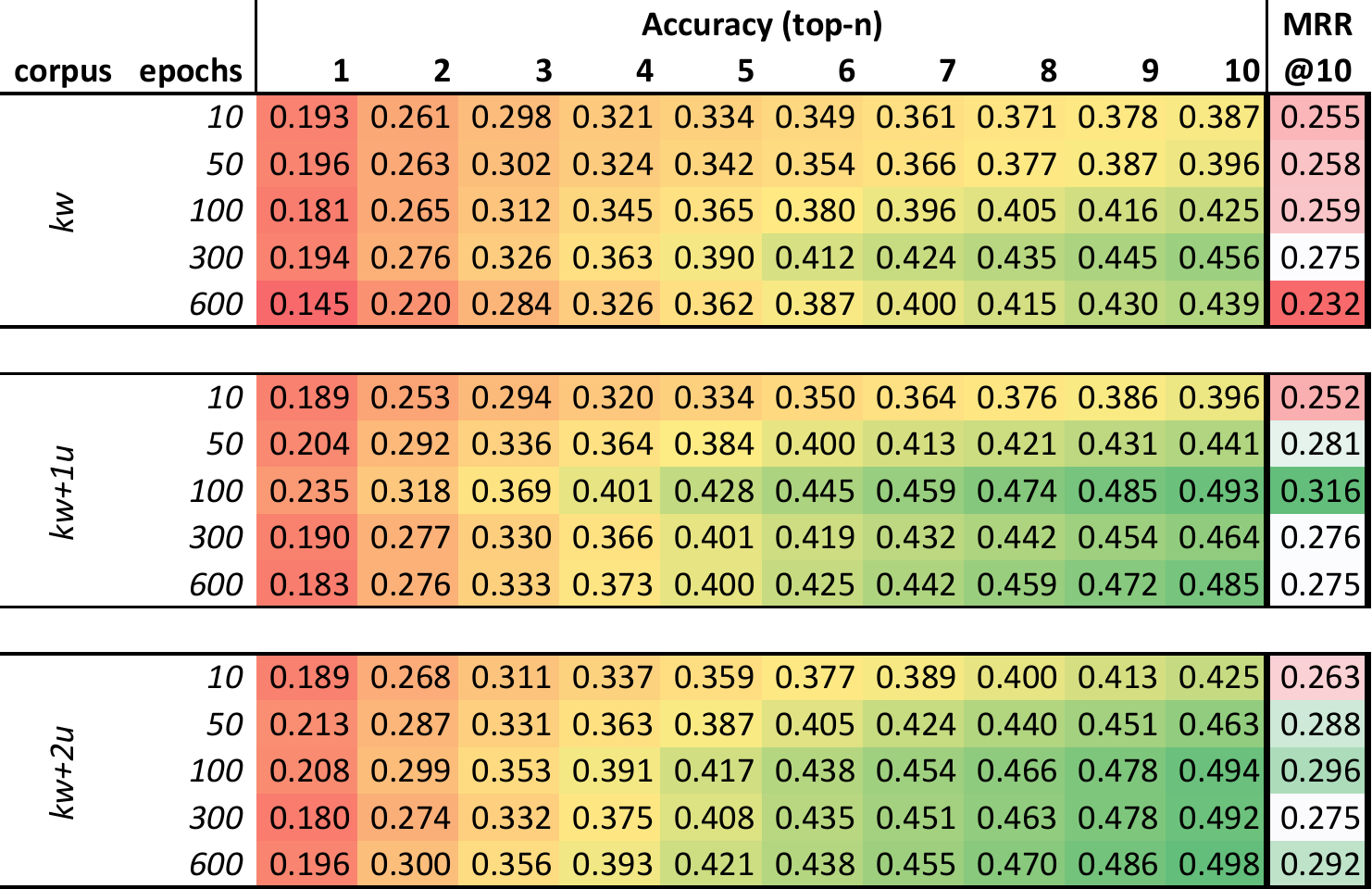}
	\caption{Performance metrics in terms of top-1 to top-10 accuracy and MRR@10 for the \emph{tensorflow\_embedding} pipeline with spell-checking for different training corpora and a different number of training \texttt{epochs}.}
	\label{fig:res_epochs}
\end{figure}

\runinhead{Results.}
Figure \ref{fig:res_pipeline} shows the accuracy and MRR values for all conditions. We only restrict \emph{tensorflow\_embedding} to the default number of \texttt{epochs} which is $300$.
At the corpus level, we can observe that the accuracy and the MRR increase when training with additional user annotations for all pipeline configurations. For example, the \emph{spacy\_sklearn} pipeline without spell-checking achieves a top-10 accuracy of $0.317$ and a MRR of $0.139$ when using the \texttt{kw} training corpus with keywords only. Both measures increase to $0.33$ and $0.165$, respectively, when adding two natural queries for training. In some cases, adding only 1 user query results in slightly better scores. However, the overall trend is that more user annotations yield better results.

In addition, we observe performance improvements for pipelines that use our spell-checking component when compared to the default pipelines that do not make use of it: The \emph{spacy\_sklearn} \texttt{kw+2u} condition performs $9.8\%$ better, the \emph{tensorflow\_embedding} \texttt{kw+2u} condition performs $3.8\%$ better, in terms of top-10 accuracy. We can observe similar improvements for the majority of included metrics.
Similar to the differentiation by corpus, we can find cases where spell-checking reduces the performance for a particular measure, against the overall trend.

Overall, the \emph{tensorflow\_embedding} pipelines perform considerably better than the \emph{spacy\_sklearn} pipeline irrespective of the remaining parameter configuration: the best performing methods are achieved by the \emph{tensorflow\_embedding} pipeline with spell-checking.
Figure \ref{fig:res_epochs} sheds more light on this particular setting. It provides performance measures for all corpora and for different number of \texttt{epochs} used for model training. Pipelines that use $300$ \texttt{epochs} for training range among the best for all corpora. When adding more natural user annotations, using $100$ \texttt{epochs} achieves similar or better scores, in particular concerning the top-10 accuracy and the MRR.
Re-ranking the top-10 results can only improve the performance in QA, if the correct answer is among the top-10 results. Therefore, we use the \emph{tensorflow\_embedding} pipeline with spellchecking, $100$ \texttt{epochs} and the full training corpus as baseline for evaluating the re-ranking approach.

\section{Re-Ranking Approach}

\begin{figure*}
	\centering
	\includegraphics[width=\textwidth]{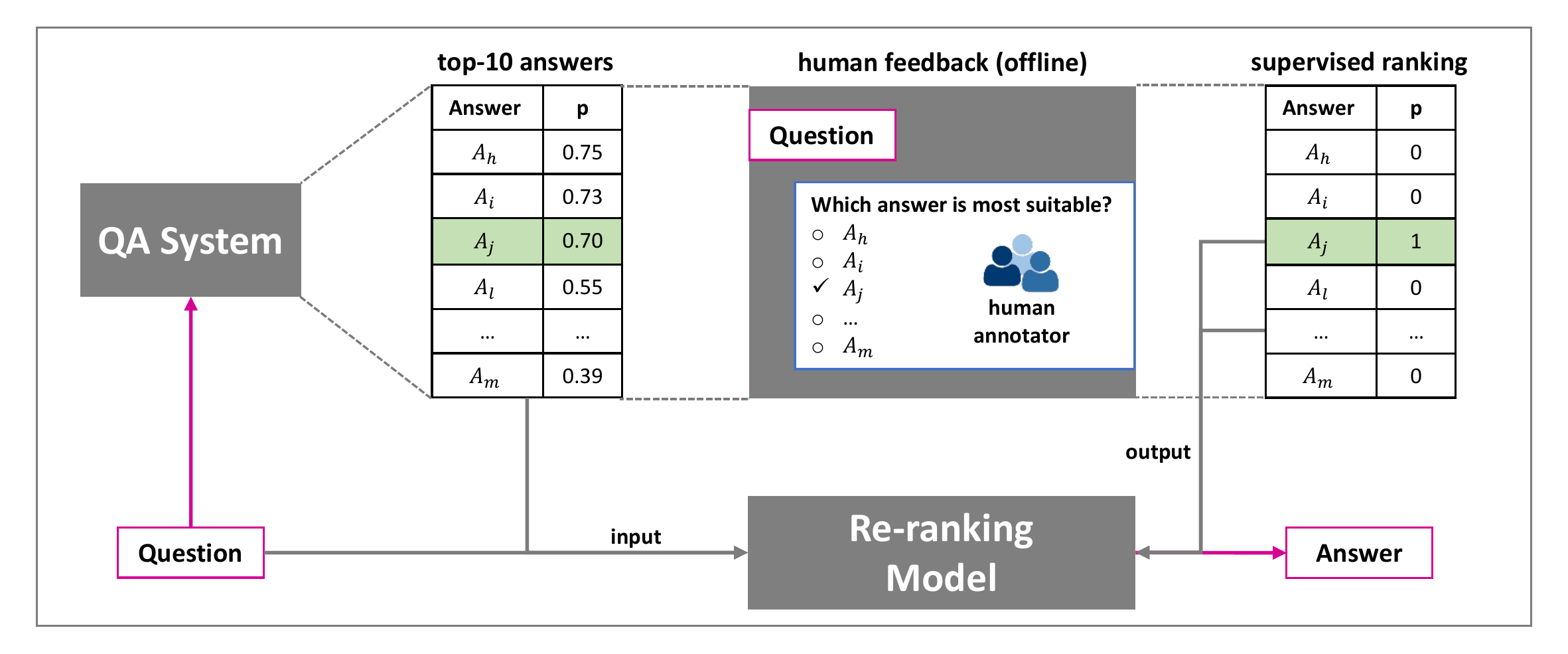}
	\caption{Complete QA system architecture including the re-ranking model. The re-ranking model is trained using manually annotated data for generating a supervised/ideal ranking result for the top-10 answers from the QA system. Features are extracted from the user question and a particular answer candidate. At inference time, the re-ranking model is used to improve the initial top-10 ranking. }
	\label{fig:reranking}
\end{figure*}

Our re-ranking approach compares a user query with the top-10 results of the baseline QA system. In contrast to the initial ranking, our re-ranking takes the content of the answer candidates into account instead of encoding the user query only. 
Our algorithm compares the text of the recent user query to each result. We include the answer text and the confidence value of the baseline system for computing a similarity estimate. Finally, we re-rank the results by their similarity to the query (see Algorithm \ref{alg:reranking}).

\begin{algorithm}
	\KwIn{a user query $q$; the corresponding list of top-10 results $R$ including an answer $a$ and the baseline confidence $c$;}
	\KwOut{an updated ranking $R'$}
	\BlankLine
	\Begin{
		$R' \longleftarrow []$\;
		\ForEach{$(c,a) \in R$}
		{
			$c' \longleftarrow similarity(q, a, c)$\;
			$R'.append((c',a))$\;
		}
		\tcp{sort R' by confidences c', descending}	
		$sort(R')$\;
		\KwRet{$R'$}
	}
	\caption{Re-Ranking Algorithm}
	\label{alg:reranking}
\end{algorithm}

We consider a data-driven similarity function that compares linguistic features of the user query and answer candidates and also takes into account the confidence of the baseline QA system. This similarity estimate shall enhance the baseline by using an extended data and feature space, but without neglecting the learned patterns of the baseline system. The possible improvement in top-1 accuracy is limited by the top-10 accuracy of the baseline system ($49.4\%$), because our re-ranking cannot choose from the remaining answers. Figure \ref{fig:reranking} shows how the re-ranking model is connected to the deployed QA system: it requires access to its in- and outputs for the additional ranking step.

We consider the gradient boosted regression tree for learning a similarity function for re-ranking similar to \cite{Savenkov2016a}. The features for model training are extracted from pre-processed query-answer pairs. Pre-processing includes tokenization and stemming of query and answer and the extraction of uni-, bi- and tri-grams from both token sequences\footnote{We use default word tokenizer, Snowball stemmer and n-gram extraction of the nltk toolkit \cite{bird2009natural}}. We include three distance metrics as feature: the Jaccard distance, the cosine similarity\footnote{We use the implementation for Jaccard distance and cosine similarity as found in the following Github gist: \href{https://gist.github.com/gaulinmp/da5825de975ed0ea6a24186434c24fe4}{gaulinmp/similarity\_example.ipynb}}, and the plain number of n-gram matches between n-grams of a query and an answer.

\begin{algorithm}
	\KwIn{a train- and test split of the evaluation corpus $C_{train},C_{test}$, each including QA-pairs as tuples $(q,a^+)$; the pre-trained baseline QA model for initial ranking $r$ and the untrained re-ranking model $r'$.}
	\KwOut{evaluation metrics.}
	\BlankLine
	\Begin{
		\tcp{training of the re-ranking model}
		$X \longleftarrow []$\;
		$y \longleftarrow []$\;
		\ForEach{$(q,a^+) \in C_{train}$}{
			$R \longleftarrow r.rank(q)$ \tcp*{R contains top-10 results}
			\If{$a^+ \notin R)$}{continue with next QA pair}
			\Else{
				\tcp{add positive sample}
				$c^+ \longleftarrow R[a^+]$ \tcp*{confidence for $a^+$}				
				$X.append(features\_from(q,a^+,c^+))$\;
				$y.append(1)$\;
				\tcp{add negative sample}
				$a^- \longleftarrow $ random $a \in R\setminus{a^+}$\;
				$c^- \longleftarrow R[a^-]$\;
				$X.append(features\_from(q,a^-, c^-))$\;
				$y.append(0)$\;
			}
		}
		$r'.train(X, y)$\;
		\BlankLine
		\tcp{evaluation of the re-ranking model}
		$results \longleftarrow \emptyset$\;
		\ForEach{$(q,a^+) \in C_{test}$}{
			$R \longleftarrow r.rank(q)$ \tcp*{top-10 baseline ranking}
			$R' \longleftarrow r'.rank(q, R)$ \tcp*{\textbf{apply re-ranking}}
			$results.append(R')$\;
		}
		\Return{$compute\_metrics(results)$}
	}
	\caption{Evaluation Procedure (per Data Split)}
	\label{alg:eval}
\end{algorithm}

\section{Re-Ranking Performance Evaluation}

We compare our data-driven QA system with a version that re-ranks resulting top-10 candidates using the additional ranking model. 
We want to answer the question whether our re-ranking approach can improve the performance of the baseline QA pipeline after deployment.
For that, we use the evaluation corpus ($n=3084$) for training and evaluating our re-ranking method using 10-fold cross-validation, i.e., $90\%$ of the data is used for training and $10\%$ for testing with $10$ different train-test splits. 

The training and testing procedure per data split of the cross-validation is shown in Algorithm \ref{alg:eval}. For each sample query $q$ in the train set $C_{train}$, we include the correct answer $a^+$ and one randomly selected negative answer candidate $a^-$ for a balanced model training. We skip a sample, if the correct answer is not contained in the top-10 results: we include $49.4\%$ of the data (see top-10 accuracy of the baseline QA model in Figure \ref{fig:res_epochs}).
The baseline QA model $r$ and the trained re-ranking method $r'$ are applied to all sample queries in the test set $C_{test}$. Considered performance metrics are computed using the re-ranked top-10 $results$.
We repeat the cross-validation $5$ times to reduce effects introduced by the random selection of negative samples. We report the average metrics from $10$ cross-validation folds and the $5$ repetitions of the evaluation procedure.

\runinhead{Results.}
The averaged cross-validation results of our evaluation, in terms of top-n accuracies and the MRR@10, are shown in Table \ref{tab:accuracy}: the top-1 to top-9 accuracies improve consistently. The relative improvement decreases from $14.83\%$ for the top-1 accuracy to $1.68\%$ for the top-9 accuracy. The top-10 accuracy stays constant, because the re-ranking cannot choose from outside the top-10 candidates. The MRR improves from $0.296$ to $0.323$ ($9.15\%$).

\begin{table}
	\centering
	\caption{Performance metrics of the baseline QA pipeline and our re-ranking method ($n=3084$).}
	\label{tab:accuracy}
	\begin{tabular}{l|ll|l}
							& \multicolumn{2}{c|}{\textbf{Method}}		& \textbf{Relative} \\
		\textbf{Metric}		& \emph{Baseline QA} & \emph{Re-Ranking}	& \textbf{Improvement} \\
		\hline
		top-1 accuracy		& $0.208$ 		& $0.239$	& $14.83\%$	\\
		top-2 accuracy		& $0.299$ 		& $0.334$	& $11.84\%$ \\
		top-3 accuracy		& $0.353$ 		& $0.384$	& $8.99\%$ \\
		top-4 accuracy		& $0.391$ 		& $0.415$	& $6.31\%$ \\
		top-5 accuracy		& $0.417$ 		& $0.44$	& $5.59\%$ \\
		top-6 accuracy		& $0.438$ 		& $0.459$ 	& $4.83\%$ \\
		top-7 accuracy		& $0.454$ 		& $0.471$	& $3.74\%$ \\
		top-8 accuracy		& $0.466$ 		& $0.48$	& $3.02\%$ \\
		top-9 accuracy		& $0.478$ 		& $0.486$	& $1.68\%$ \\
		top-10 accuracy		& $0.494$		& $0.494$	& $0.00\%$ \\
		\hline
		MRR@10 				& $0.296$		& $0.323$	& $9.15\%$ \\
		\hline
	\end{tabular}
\end{table}

\section{Discussion}

Our results indicate that the accuracy of the described QA system benefits from our re-ranking approach. Hence, it can be applied to improve the performance of already deployed QA systems that provide a top-10 ranking with confidences as output.
However, the performance gain is small, which might have several reasons. For example, we did not integrate spell-checking in our re-ranking method which proved to be effective in our baseline evaluation. Further, the re-ranking model is based on very simple features. It would be interesting to investigate the impact of more advanced features, or models, on the ranking performance (e.g., word embeddings \cite{Mikolov2013} and deep neural networks for learning similarity functions \cite{Das2016,Minaee2017}).
Nevertheless, as can be seen in examples 1, 2 and 4 in Table \ref{tab:qa_samples}, high-ranked but incorrect answers are often meaningful with respect to the query: the setting in our evaluation is overcritical, because we count incorrect, but meaningful answers as negative result.
A major limitation is that the re-ranking algorithm cannot choose answer candidates beyond the top-10 results. It would be interesting to classify whether an answer is present in the top-10 or not. If not, the algorithm could search outside the top-10 results. Such a meta-model can also be used to estimate weaknesses of the QA model: it can determine topics that regularly fail, for instance, to guide data labelling for a targeted improvement of the model, also known as active learning \cite{Settles2010}, and in combination with techniques from semi-supervised learning \cite{N18-2092,Chang2016}.

Data labelling and incremental model improvement can be scaled by crowdsourcing. Examples include the parallel supervision of re-ranking results and targeted model improvement as human oracles in an active learning setting.
Results from crowd-supervised re-ranking allows us to train improved re-ranking models \cite{Savenkov2016a,Huang2018}, but also a meta-model that detects queries which are prone to error.
The logs of a deployed chatbot, that contain actual user queries, can be efficiently analysed using such a meta-model to guide the sample selection for costly human data augmentation and creation. An example of a crowdsourcing approach that could be applied to our QA system and data, with search logs can be found in \cite{Bernstein2012}.


\section{Conclusion}
We implemented a simple re-ranking method and showed that it can effectively improve the performance of QA systems after deployment.
Our approach includes the top-10 answer candidates and confidences of the initial ranking for selecting better answers.
Promising directions for future work include the investigation of more advanced ranking approaches for increasing the performance gain and continuous improvements through crowdsourcing and active learning.

\bibliographystyle{plain}
\bibliography{references}

\begin{thebibliography}{10}

\bibitem{tensorflow2015-whitepaper}
Mart\'{\i}n Abadi, Ashish Agarwal, Paul Barham, Eugene Brevdo, Zhifeng Chen,
  Craig Citro, Greg~S. Corrado, Andy Davis, Jeffrey Dean, Matthieu Devin,
  Sanjay Ghemawat, Ian Goodfellow, Andrew Harp, Geoffrey Irving, Michael Isard,
  Yangqing Jia, Rafal Jozefowicz, Lukasz Kaiser, Manjunath Kudlur, Josh
  Levenberg, Dandelion Man\'{e}, Rajat Monga, Sherry Moore, Derek Murray, Chris
  Olah, Mike Schuster, Jonathon Shlens, Benoit Steiner, Ilya Sutskever, Kunal
  Talwar, Paul Tucker, Vincent Vanhoucke, Vijay Vasudevan, Fernanda Vi\'{e}gas,
  Oriol Vinyals, Pete Warden, Martin Wattenberg, Martin Wicke, Yuan Yu, and
  Xiaoqiang Zheng.
\newblock {TensorFlow}: Large-scale machine learning on heterogeneous systems,
  2015.
\newblock Software available from tensorflow.org.

\bibitem{Bernstein2012}
Michael~S. Bernstein, Jaime Teevan, Susan Dumais, Daniel Liebling, and Eric
  Horvitz.
\newblock {Direct answers for search queries in the long tail}.
\newblock In {\em Proceedings of the 2012 ACM annual conference on Human
  Factors in Computing Systems - CHI '12}, page 237, New York, New York, USA,
  2012. ACM Press.

\bibitem{bird2009natural}
Steven Bird, Ewan Klein, and Edward Loper.
\newblock {\em Natural language processing with Python: analyzing text with the
  natural language toolkit}.
\newblock " O'Reilly Media, Inc.", 2009.

\bibitem{Bocklisch2017}
Tom Bocklisch, Joey Faulkner, Nick Pawlowski, and Alan Nichol.
\newblock {Rasa: Open Source Language Understanding and Dialogue Management}.
\newblock 12 2017.

\bibitem{DBLP:journals/corr/BuckBCGHGW17}
Christian Buck, Jannis Bulian, Massimiliano Ciaramita, Andrea Gesmundo, Neil
  Houlsby, Wojciech Gajewski, and Wei Wang.
\newblock {Ask the Right Questions: Active Question Reformulation with
  Reinforcement Learning}.
\newblock {\em CoRR}, abs/1705.0, 2017.

\bibitem{Chang2016}
Joseph~Chee Chang, Aniket Kittur, and Nathan Hahn.
\newblock {Alloy: Clustering with Crowds and Computation}.
\newblock In {\em Proceedings of the 2016 CHI Conference on Human Factors in
  Computing Systems - CHI '16}, pages 3180--3191, New York, New York, USA,
  2016. ACM Press.

\bibitem{P17-1171}
Danqi Chen, Adam Fisch, Jason Weston, and Antoine Bordes.
\newblock Reading wikipedia to answer open-domain questions.
\newblock In {\em Proceedings of the 55th Annual Meeting of the Association for
  Computational Linguistics (Volume 1: Long Papers)}, pages 1870--1879. ACL,
  2017.

\bibitem{Das2016}
Arpita Das, Harish Yenala, Manoj Chinnakotla, and Manish Shrivastava.
\newblock {Together we stand: Siamese Networks for Similar Question Retrieval}.
\newblock In {\em Proceedings of the 54th Annual Meeting of the Association for
  Computational Linguistics (Volume 1: Long Papers)}, volume~1, pages 378--387,
  Stroudsburg, PA, USA, 2016. Association for Computational Linguistics.

\bibitem{N18-2092}
Bhuwan Dhingra, Danish Danish, and Dheeraj Rajagopal.
\newblock {Simple and Effective Semi-Supervised Question Answering}.
\newblock In {\em Proceedings of the 2018 Conference of the North American
  Chapter of the Association for Computational Linguistics: Human Language
  Technologies, Volume 2 (Short Papers)}, pages 582--587. ACL, 2018.

\bibitem{eric-manning:2017:SIGDIAL}
Mihail Eric and Christopher~D Manning.
\newblock {Key-Value Retrieval Networks for Task-Oriented Dialogue}.
\newblock In {\em Proceedings of the 18th Annual SIGdial Meeting on Discourse
  and Dialogue}, pages 37--49, Saarbr{\"{u}}cken, Germany, 8 2017. Association
  for Computational Linguistics.

\bibitem{Huang2018}
Ting-Hao~'Kenneth' Huang, Joseph~Chee Chang, and Jeffrey~P. Bigham.
\newblock {Evorus: A Crowd-powered Conversational Assistant Built to Automate
  Itself Over Time}.
\newblock 1 2018.

\bibitem{Huang2015}
Ting-Hao~Kenneth Huang, Walter~S. Lasecki, and Jeffrey~P. Bigham.
\newblock {Guardian: A Crowd-Powered Spoken Dialog System for Web APIs}.
\newblock In {\em Third AAAI Conference on Human Computation and
  Crowdsourcing}, 2015.

\bibitem{Joshi2017}
Mandar Joshi, Eunsol Choi, Daniel~S Weld, and Luke Zettlemoyer.
\newblock {TriviaQA: {A} Large Scale Distantly Supervised Challenge Dataset for
  Reading Comprehension}.
\newblock In Regina Barzilay and Min-Yen Kan, editors, {\em Proceedings of the
  55th Annual Meeting of the Association for Computational Linguistics, {ACL}
  2017, Volume 1: Long Papers}, pages 1601--1611. ACL, 2017.

\bibitem{Mikolov2013}
Tomas Mikolov, Kai Chen, Greg Corrado, and Jeffrey Dean.
\newblock {Efficient Estimation of Word Representations in Vector Space}.
\newblock 1 2013.

\bibitem{Minaee2017}
Shervin Minaee and Zhu Liu.
\newblock {Automatic Question-Answering Using A Deep Similarity Neural
  Network}.
\newblock 8 2017.

\bibitem{Oraby2017}
Shereen Oraby, Pritam Gundecha, Jalal Mahmud, Mansurul Bhuiyan, and Rama
  Akkiraju.
\newblock {"How May I Help You?": Modeling Twitter Customer
  ServiceConversations Using Fine-Grained Dialogue Acts}.
\newblock In {\em Proceedings of the 22nd International Conference on
  Intelligent User Interfaces - IUI '17}, pages 343--355, New York, New York,
  USA, 2017. ACM Press.

\bibitem{scikit-learn}
F~Pedregosa, G~Varoquaux, A~Gramfort, V~Michel, B~Thirion, O~Grisel, M~Blondel,
  P~Prettenhofer, R~Weiss, V~Dubourg, J~Vanderplas, A~Passos, D~Cournapeau,
  M~Brucher, M~Perrot, and E~Duchesnay.
\newblock {Scikit-learn: Machine Learning in Python}.
\newblock {\em Journal of Machine Learning Research}, 12:2825--2830, 2011.

\bibitem{D16-1264}
Pranav Rajpurkar, Jian Zhang, Konstantin Lopyrev, and Percy Liang.
\newblock Squad: 100,000+ questions for machine comprehension of text.
\newblock In {\em Proceedings of the 2016 Conference on Empirical Methods in
  Natural Language Processing}, pages 2383--2392. ACL, 2016.

\bibitem{Savenkov2016a}
Denis Savenkov and Eugene Agichtein.
\newblock {CRQA: Crowd-Powered Real-Time Automatic Question Answering System}.
\newblock {\em Fourth AAAI Conference on Human Computation and Crowdsourcing},
  2016.

\bibitem{Serban2015a}
Iulian~V. Serban, Alessandro Sordoni, Yoshua Bengio, Aaron Courville, and
  Joelle Pineau.
\newblock {Building End-To-End Dialogue Systems Using Generative Hierarchical
  Neural Network Models}.
\newblock 7 2015.

\bibitem{Serban2015}
Iulian~Vlad Serban, Ryan Lowe, Peter Henderson, Laurent Charlin, and Joelle
  Pineau.
\newblock {A Survey of Available Corpora for Building Data-Driven Dialogue
  Systems}.
\newblock 12 2015.

\bibitem{Settles2010}
Burr Settles.
\newblock {Active learning literature survey}.
\newblock {\em University of Wisconsin, Madison}, 52(55-66):11, 2010.

\bibitem{Sonntag2008}
Daniel Sonntag.
\newblock {On Introspection, Metacognitive Control and Augmented Data Mining
  Live Cycles}.
\newblock jul 2008.

\bibitem{Sonntag09}
Daniel Sonntag.
\newblock Introspection and adaptable model integration for dialogue-based
  question answering.
\newblock In {\em {IJCAI}}, pages 1549--1554, 2009.

\bibitem{Sonntag2010_book}
Daniel Sonntag.
\newblock {\em {Ontologies and Adaptivity in Dialogue for Question Answering}},
  volume~4 of {\em Studies on the Semantic Web}.
\newblock AKA and IOS Press, Heidelberg, first edition, 2010.

\bibitem{SonntagEHPPRR07}
Daniel Sonntag, Ralf Engel, Gerd Herzog, Alexander Pfalzgraf, Norbert Pfleger,
  Massimo Romanelli, and Norbert Reithinger.
\newblock Smartweb handheld - multimodal interaction with ontological knowledge
  bases and semantic web services.
\newblock In {\em Artifical Intelligence for Human Computing, {ICMI} 2006 and
  {IJCAI} 2007 International Workshops, Banff, Canada, November 3, 2006,
  Hyderabad, India, January 6, 2007, Revised Seleced and Invited Papers}, pages
  272--295, 2007.

\bibitem{Williams2017}
Jason~D Williams, Kavosh Asadi, and Geoffrey Zweig.
\newblock {Hybrid Code Networks: practical and efficient end-to-end dialog
  control with supervised and reinforcement learning}.
\newblock In {\em Proceedings of the 55th Annual Meeting of the Association for
  Computational Linguistics (Volume 1: Long Papers)}, volume~1, pages 665--677,
  Stroudsburg, PA, USA, 2017. Association for Computational Linguistics.

\bibitem{DBLP:journals/corr/abs-1709-03856}
Ledell Wu, Adam Fisch, Sumit Chopra, Keith Adams, Antoine Bordes, and Jason
  Weston.
\newblock Starspace: Embed all the things!
\newblock {\em CoRR}, abs/1709.03856, 2017.

\bibitem{Xue2008}
Xiaobing Xue, Jiwoon Jeon, and W.~Bruce Croft.
\newblock {Retrieval models for question and answer archives}.
\newblock In {\em Proceedings of the 31st annual international ACM SIGIR
  conference on Research and development in information retrieval - SIGIR '08},
  page 475, New York, New York, USA, 2008. ACM Press.

\end{thebibliography}

\end{document}